\documentclass[letterpaper, 10pt, conference]{ieeeconf}

\DeclareRobustCommand{\rchi}{{\mathpalette\irchi\relax}}
\newcommand{\irchi}[2]{\raisebox{\depth}{$#1\chi$}} 

\IEEEoverridecommandlockouts
\usepackage{xcolor}
\usepackage{amsmath,amssymb,stmaryrd}   
\usepackage{centernot}
\usepackage{mathtools}
\usepackage{stmaryrd}

\makeatletter
\newcommand{\xMapsto}[2][]{\ext@arrow 0599{\Mapstofill@}{#1}{#2}}
\def\Mapstofill@{\arrowfill@{\Mapstochar\Relbar}\Relbar\Rightarrow}
\makeatother

\usepackage{xcolor,soul,framed}
\usepackage{tabularx}
\usepackage{todonotes}
\colorlet{shadecolor}{yellow}

\usepackage{graphicx}
\usepackage{caption}
\usepackage{subfigure}
\graphicspath{{../pdf/}{../jpeg/}}
\DeclareGraphicsExtensions{.pdf,.jpeg,.png}

\newcommand{\comment}[1]{}

\begin{document}

\title{\LARGE \bf Predicting Future Occupancy Grids in Dynamic Environment with Spatio-Temporal Learning}

\author{Khushdeep S. Mann$^1$, Abhishek Tomy$^1$, Anshul Paigwar$^1$, Alessandro Renzaglia$^2$ and Christian Laugier$^1$ 
      
\thanks{$^1$ University of Grenoble Alpes, Inria, 38000, Grenoble, France;
e-mail: {\tt\small firstname.lastname@inria.fr}}
\thanks{$^2$ Univ Lyon, Inria, INSA Lyon, CITI, F-69621 Villeurbanne, France; e-mail: {\tt\small firstname.lastname@inria.fr}}
  }

\maketitle
\thispagestyle{empty}
\pagestyle{empty}

\begin{abstract}

Reliably predicting future occupancy of highly dynamic urban environments is an important precursor for safe autonomous navigation. Common challenges in the prediction include forecasting the relative position of other vehicles, modelling the dynamics of vehicles subjected to different traffic conditions, and vanishing surrounding objects. To tackle these challenges, we propose a spatio-temporal prediction network pipeline that takes the past information from the environment and semantic labels separately for generating future occupancy predictions. Compared to the current SOTA, our approach predicts occupancy for a longer horizon of 3 seconds and in a relatively complex environment from the nuScenes dataset. Our experimental results demonstrate the ability of spatio-temporal networks to understand scene dynamics without the need for HD-Maps and explicit modeling dynamic objects. We publicly release our occupancy grid dataset based on nuScenes to support further research.
\end{abstract}



\section{Introduction}

The ability to safely and intelligently navigate through different traffic scenarios is an important aspect of Autonomous Vehicles (AVs). Such navigation is relatively easy for human drivers because they can interpret and predict how the surrounding environment will evolve. However, performing a similar prediction task is quite difficult for AVs especially under different traffic conditions. Predictions in the 3D world coordinates require the entire pipeline starting from the detection of the 3D object in the LiDAR point cloud along with the association of agents from the previous frames to finally performing predictions using the previous sequence to work accurately. Through this prediction pipeline, error can creep in from any of the modules and can affect the final results.

In this paper, we tackle the challenge of performing long-term future predictions for AVs with Occupancy Grid Maps (OGMs). OGMs are widely used in the industry as they capture the local environment and dynamic agents into discrete spatial cells. Each cell is classified as static, dynamic, unknown, or free space. OGMs are probabilistic representations, integrating uncertainties from sensors, and temporal analysis. They provide a dense representation as each portion of the space is analyzed and not limited to objects of a given type or size.

\begin{figure}[!t]
	\centering
		\includegraphics[width=.5\textwidth]{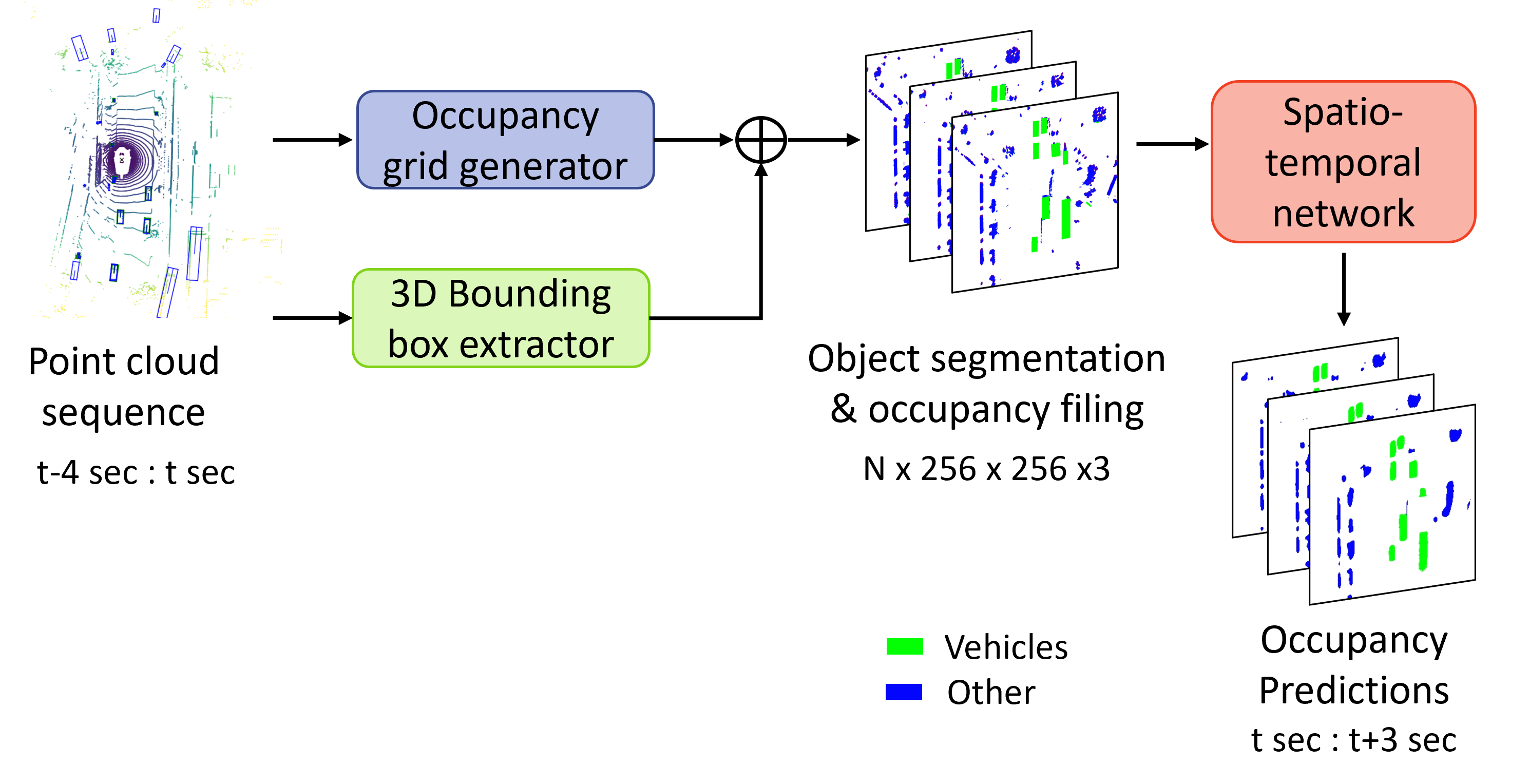}
	\caption{Generated OGMs from nuScenes dataset are coupled with corresponding ground truth 3D labels and fed into an spatio-temporal prediction network pipeline. Network learns from past 4 sec contextual information and predicts 3 sec into the future. }
	\label{fig:teaser}
\end{figure}

Future frame prediction of the OGMs is equivalent to a video prediction task with certain domain-specific knowledge and modifications in the design of the model to capture the discretized spatial information of agents and environments in a normal driving scenario. Accordingly, recent works have considered deep learning-based models for predicting future video frames with OGMs \cite{ Mohajerin2019MultiStepPO,  itkina2019dynamic}. The existing works usually separate the static and dynamic part for prediction \cite{toyungyernsub2020double}. However, in our work, we propose a semantic scheme that can better represent the different dynamics of specific agents compared to a dynamic cell that may contain multiple agents such as pedestrians, cyclists or a vehicle. 

Main contributions of this work are: 
\begin{itemize}

    \item A Spatio-Temporal Network Pipeline for long-term future occupancy grid prediction. Our approach uses semantic labels of the vehicle in OGMs to model the specific motion type of agents in the prediction rather than using a generic combined prediction of static and dynamic cells that contains the environment and various types of agents such as cars, pedestrians, cyclists, etc. 
    
    \item Publicly releasing an OGMs dataset consisting of static environment and semantic labels for ease in long-term prediction. Presenting the results for this dataset with state-of-art video prediction models shows that under this training scheme the models have consistent performance even for long-term predictions. The code and dataset are available at \cite{dataset}.
    
    \item Evaluate the effectiveness of a spatio-temporal module in comparison to ConvLSTM for modelling the spatial and temporal nature of future prediction in OGMs.
\end{itemize}

\section{Related work}

\subsection{ Video prediction task}

In recent years, deep learning based video prediction has emerged as a promising research direction with applications in future sequence prediction to restore a high-resolution video from its corresponding low-resolution frames. Video prediction is defined as a self-supervised learning task to extract meaningful temporal and spatial patterns in natural videos \cite{oprea2020review}. Within the video-prediction framework, sequence-to-sequence architecture that takes in a sequence of images and predicts future frames is a suitable model for our task. Developments in the video-prediction task have been adapted in the autonomous vehicle domain to predict future trajectories of agents and RGB frames to predict future occupancy grids. In the sequence-to-sequence architectures, CNNs are used to model spatial information, while RNN is used to incorporate the temporal information from the sequence. The combination of an CNN and RNN can handle the dynamic agents in the environment and extract the necessary spatial relations for per-pixel prediction of both static and dynamic objects in the environment \cite{zhou2020deep}.

In this work, we compare the performance of two video prediction approaches, a ConvLSTM \cite{xingjian2015convolutional} and a PredRNN \cite{wang2021predrnn} based network architecture, over the OGMs dataset. ConvLSTM is based on prediction networks for language and speech modelling tasks. In this, LSTM is used to preserve the long-term temporal dynamics and non-markovian properties in a network structure called a memory cell and the convolutional layers act as filters to extract important features from images. PredRNN builds upon this to include spatial correlation and temporal dynamics in future image prediction with the help of novel Spatio-Temporal LSTM (ST-LSTM) blocks. In PredRNN, a new memory bank is added that takes as input the memory flow from the output of the previous frame ST-LSTM layer in the sequence. This enables the module to learn complex spatial relations of the transition between consecutive frames.

\subsection{ Occupancy grid prediction}

Path prediction in the occupancy grid offers the benefit of reducing the computational complexity of path association through tracking of individual subjects and prediction of individual agents in large and cluttered driving scenarios. Also, the interaction with other agents and the static environment is implicitly captured by the occupancy grid prediction models. The first end-to-end object tracking approach to directly map raw sensor input to track objects in an indoor environment with occlusion was presented in \cite{ondruska2016deep}. Building upon this, authors proposed an end-to-end trainable framework that was able to track a range of objects, including buses, cars, cyclists and pedestrians through occlusion by predicting a fully non-occluded occupancy grid from raw laser input recorded from a stationary or moving vehicle with RNN for capturing the temporal evolution of the state of the environment \cite{dequaire2016deep} \cite{dequaire2018deep}.

\begin{figure}[t]
  \subfigure[]{
    \includegraphics[width=.17\textwidth]{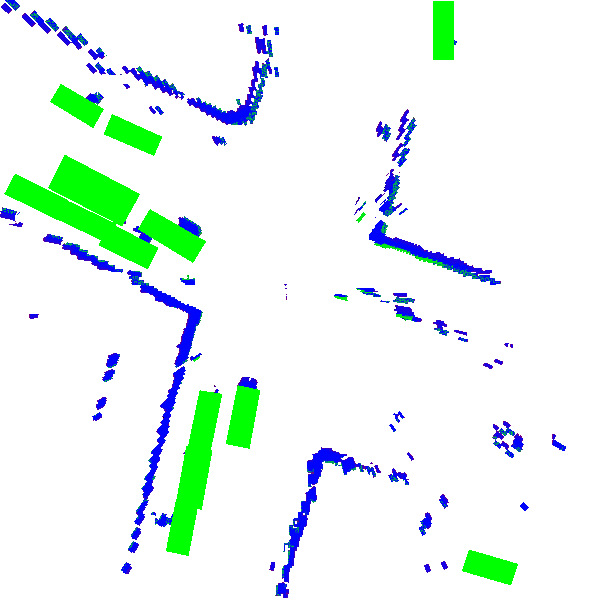}}
  \subfigure[]{
    \includegraphics[width=.17\textwidth]{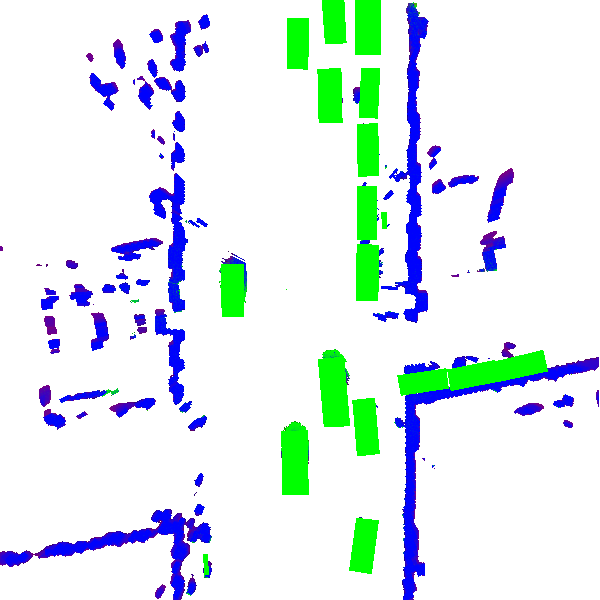}}
  \subfigure[]{
    \includegraphics[width=.17\textwidth]{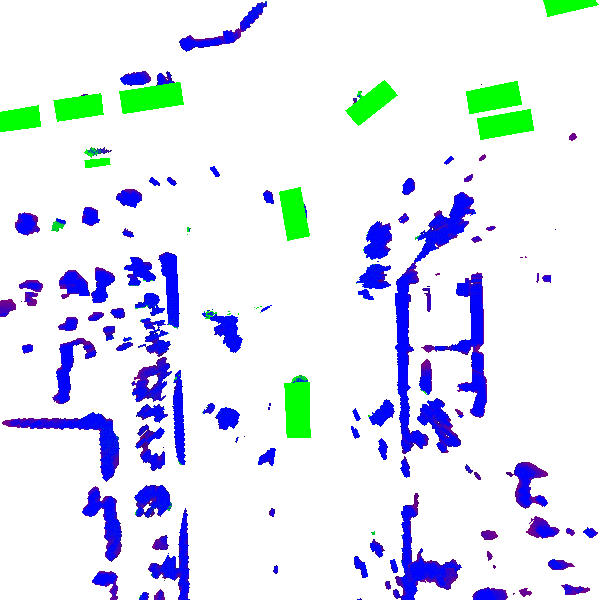} }
  \subfigure[]{
    \includegraphics[width=.17\textwidth]{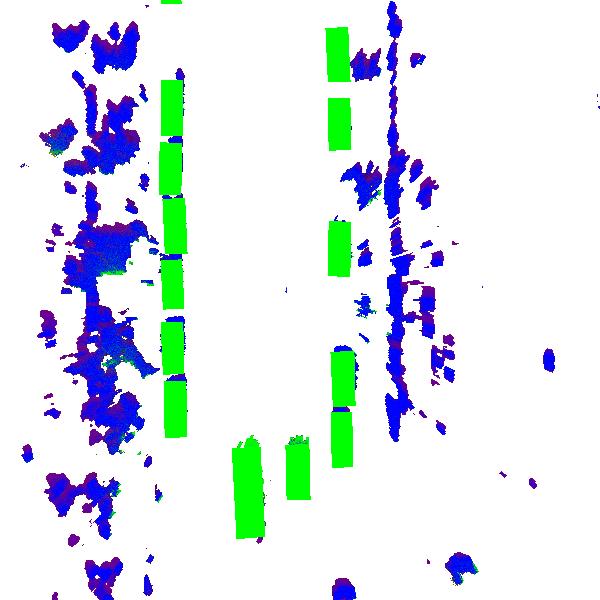} }
    \centering
    \caption{Occupancy grid maps generated from the nuScenes dataset. Agents belonging to the 'Vehicles' category are of interest and being marked by 'green' semantic pixel labels using the projections of ground truth 3D bounding boxes. Objects of any other type including the static environment are marked in 'blue'. Different road crossing motion scenarios from the dataset are presented.} 

	\label{fig:OGMs_dataset}
\end{figure}

\begin{figure*}[t]
   \centering
   {\includegraphics[width = \textwidth, height = 0.25\textwidth]{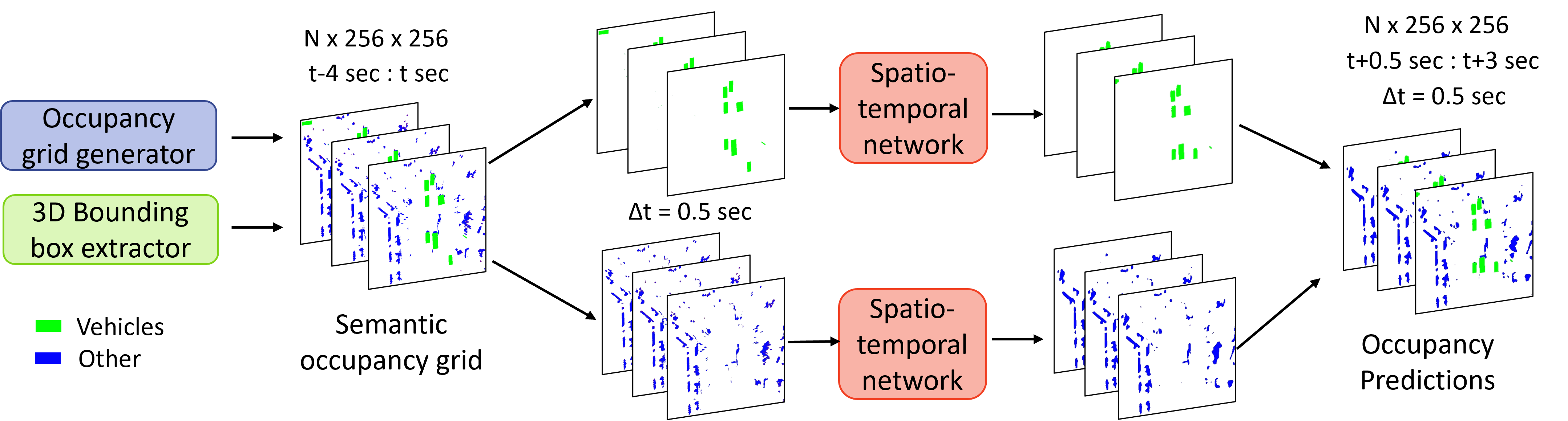}}
   \caption{Semantic occupancy grids consisting of environment and vehicles over the time frames of 0.5 sec. Grids are converted into binary images and separately fed to spatio-temporal networks. We evaluate two spatio-temporal networks: PredRNN and ConvLSTM.}
   \label{fig:model_architecture}
\end{figure*}

A multi-step prediction using RNN was proposed in \cite{Mohajerin2019MultiStepPO} so that an accurate prediction of the drivable space was extracted for efficient planning and navigation. The method uses motion-related features between consecutive frames to enhance the prediction of dynamic objects to account for the dataset bias towards static objects. In subsequent studies, separate predictions of the static and dynamic cells were explored. A double-prong network architecture with a separate stream for predicting static and dynamic objects was proposed in \cite{toyungyernsub2020double} to retain dynamic objects and reduce blurriness in long-term prediction. In \cite{schreiber2019long}, the static and dynamic cells are predicted separately, however, the input to the network is the entire OGM along with the velocity of each cell obtained through DOGMA \cite{hoermann2018dynamic}. The proposed model uses an encoder-decoder type architecture with ConvLSTM modules along with skip connections to extract Spatio-temporal correlations.

\section{Dataset}

\subsection{Generating occupancy grid maps}
\label{generated_OGMs} 

NuScenes dataset \cite{caesar2020nuscenes} contains recordings of 1000 scenes recorded in Boston and Singapore. Each scene is 20s long and collected at 20Hz. Annotations with 3D bounding boxes are provided at 2Hz for 23 classes. NuScenes dataset contains recordings from a camera, radar, and LiDAR for environment perception along with high-quality manually annotated vectorized maps with GPU/IMU measurements. A large number of annotations and environment maps along with track id's enable research on multiple tasks such as object detection, tracking, and behavior modeling.  

We generate the OGMs from NuScenes LiDAR point clouds by considering a distance of 30m in cardinal directions from the centric ego-vehicle. We followed this scheme with regards to different traffic scenarios and to capture as many dynamic objects approaching towards or receding from the ego-vehicle from all directions. To obtain the OGMs, we adopt the Conditional Monte-Carlo Dense Occupancy Tracker (CMCDOT) \cite{rummelhard2015conditional} with a resolution of 0.1m, that eventually generates a $600\times600$ grid size. This is a spatial occupancy tracker that infers the dynamic of a scene by discrete classification of pixels as static, dynamic, unknown, or free space. Accordingly, we eliminate the unknown and free occupancies and project the ground truth 3D boxes onto the OGMs to obtain our experimental dataset shown in Figure \ref{fig:OGMs_dataset}.

\subsection{3D to 2D projection of ground truth bounding boxes over nuScenes dataset}
Existing occupancy grid generators lack semantic information about the detected objects in the environment. However, the nuScenes dataset contains annotation of the 3D bounding boxes of objects in the LiDAR point cloud at 2Hz which can be utilized as semantic information in the occupancy grid. The occupancy grids are generated with a resolution such that each pixel in the occupancy grid map corresponds to 0.1m in the LiDAR point cloud. Corresponding to annotated frames in the dataset, the generated grids are matched using their timestamps, and the top corners of the bounding box of object class `vehicle' are transformed and plotted to the grid.


\section{Model Architecture}
We consider spatio-temporal learning networks for performing the prediction task. Accordingly, we implemented state-of-art PredRNN architecture\cite{wang2021predrnn}. We use the ST-LSTM building blocks for PredRNN and include ConvLSTM architecture for baseline comparison. ST-LSTM unit is represented by ($\rchi_{t}$, \textit{$H^{l}_{t-1}$}, \textit{$C^{l}_{t-1}$}, \textit{$M^{l-1}_{t}$}), where \textit{l} $\in$ \{1, . . , \textit{L}\}  and \textit{L} is the maximum number of recurrent units. At a time instant \textit{t}, $\rchi_{t}$ is the input state, \textit{$H^{l}_{t-1}$} is the output hidden state from the previous unit, \textit{$C^{l}_{t-1}$} is the memory state and \textit{$M^{l-1}_{t}$} is the saptio-temporal memory bank jointly maintained by all nodes for learning unified representations in the recurrent network. The inclusion of this memory bank is the key difference as compared to ConvLSTM unit. We consider four recurrent ST-LSTM and standard LSTM units (\textit{L}=4) in PredRNN and ConvLSTM networks respectively. Each hidden state has 64 channels. Past time frames from \textit{t} - 4s to \textit{t}, where \textit{t} indicates the current time, over the interval of \textit{$\Delta$t} = 0.5s are considered as inputs and future predictions are done for \textit{t} + 0.5s to \textit{t} + 3s considering the same time interval as demonstrated in Figure \ref{fig:model_architecture}. 

We adopt the reverse scheduled sampling scheme (RSS) for the inputs during training the networks. Introduced in \cite{wang2021predrnn}, RSS is a curriculum learning strategy that forces sequence-to-sequence models for learning long-term dynamics. A gradual change in the encoder training process accomplishes this by replacing previously generated frames with previous ground truth frames. Following the previous work \cite{denton2018stochastic, finn2016unsupervised}, we denote a unified predictive model \textit{f$_\theta$(.)} for sequence encoding. We denote the input context frames up to time \textit{T}  as \textit{$ \rchi_{input}$} = \{\textit{$ \rchi_1$}, . . ,\textit{$ \rchi_T$}\} and predicted time frames under the RSS scheme are given by:
\begin{equation}{\label{eq:rss_equation}}
    \hat{\rchi}_{t+1} = f_\theta\left( \hat{\rchi}_{t} \xmapsto{RSS} \textit{$\rchi_t$}, \textit{$H_{t-1}$}, \textit{$C_{t-1}$}, \textit{$M_{t-1}$}\right) \quad \text{for}  \quad \textit{t} \leq  \textit{T} 
\end{equation}
where, $ \xmapsto{RSS}$ denotes the gradual change in training process by considering ground truth frame \textit{$\rchi_t$} instead of generated frame $\hat{\rchi}_{t}$. Under the RSS scheme, there exists a probability $ \epsilon_k$ $ \in$ [0,1] for sampling true frames \textit{$\rchi_t$} $ \in$ \textit{$\rchi_{input}$} or a probability (1 - $ \epsilon_k$) of sampling $ \hat{\rchi}_{t}$. $\epsilon_k$ changes with number of training iterations \textit{k}, having an initial and end values as $\epsilon_s$ and $\epsilon_e$ respectively. We formulate this increase with an exponential equation as: 
\begin{equation}{\label{eq:epsilon_enq}}
     \epsilon_k = \epsilon_e - (\epsilon_e -  \epsilon_s) \times exp\left(-\frac{k}{\alpha_e}\right) 
\end{equation}

Our model architecture consists of two training paths to explicitly learn the dynamics of the static environment and semantic vehicle objects as visible in Figure \ref{fig:model_architecture}. Accordingly, the overall loss function \textit{L}$_o$ is a weighted sum of static environment loss \textit{L}$_{static}$ and semantic loss \textit{L}$_{semantic}$: 
\begin{equation}{\label{eq:overall_loss}}
    L_{0} = L_{static} + k_{0} L_{semantic} 
\end{equation}
where \textit{k}$_o$ is the weight factor for balancing the two loss terms. The static and semantic loss functions
\begin{equation}{\label{eq:static_loss}}
    L_{static}  = \frac{1}{W \times H} \sum_{c=1}^{W \times H} |y_{static}^{*}(c) - y_{static}(c)|
\end{equation}

\begin{equation}{\label{eq:semantic_loss}}
    L_{semantic}  = \frac{1}{W \times H} \sum_{c=1}^{W \times H} \lambda_{c} (y_{semantic}^{*}(c) - y_{semantic}(c))^{2}
\end{equation}
are the mean of absolute differences of each cell \textit{c} between the ground truth labels \textit{y$^*(c)$} and predicted labels \textit{y(c)}. $\lambda_{c}$ is a weighting factor for balancing the semantic cells against the background. For each cell this factor is given by:
\begin{equation}{\label{eq:lambda}}
    \lambda_{c}  = 1 + k_{s}*y_{semantic}^{*}(c)
\end{equation}
such that $\lambda_{c}=1+k_{s}$ for semantic cells and $\lambda_{c}=1$ elsewhere. 
\section{Training and Experimentation}
The experimental dataset is described in section \ref{generated_OGMs}. We consider 80\% of the dataset for training and 20\% for testing. Original OGMs provided in  $\mathbb{R}^{600\times600\times3}$ are reduced in spatial size to $\mathbb{R}^{256\times256\times1}$. The convolutional kernel size is set to $5\times5$ inside ST-LSTM and LSTM units. Networks are trained using the Adam optimizer \cite{kingma2014adam} with a starting learning rate of $3\times10^{-4}$ with a batch size of 8 sequences. We set $\epsilon_e$ = 1.0, $\epsilon_s$ = 0.5, and $\alpha_e$ = $5\times10^3$ in equation (\ref{eq:epsilon_enq}). We perform training for 25 epochs on 6109 sequences. The models are tested on 1200 test sequences considering the same input and prediction scheme. We set \textit{k}$_o$ = 10 in eqn. (\ref{eq:overall_loss}) and \textit{k}$_s$ = 2 in eqn. (\ref{eq:lambda}) based on the validation performance. We also compare the results of semantic MSE between the spatio-temporal networks and linear projection model \cite{patil:hal-03416222}. Linear projection assumes constant instantaneous velocity and uses it to evaluate the future position.


\section{Results}
\subsection{Qualitative results}

\begin{figure}[!t]
  \subfigure[Ground truth]{
    \includegraphics[width=.475\textwidth]{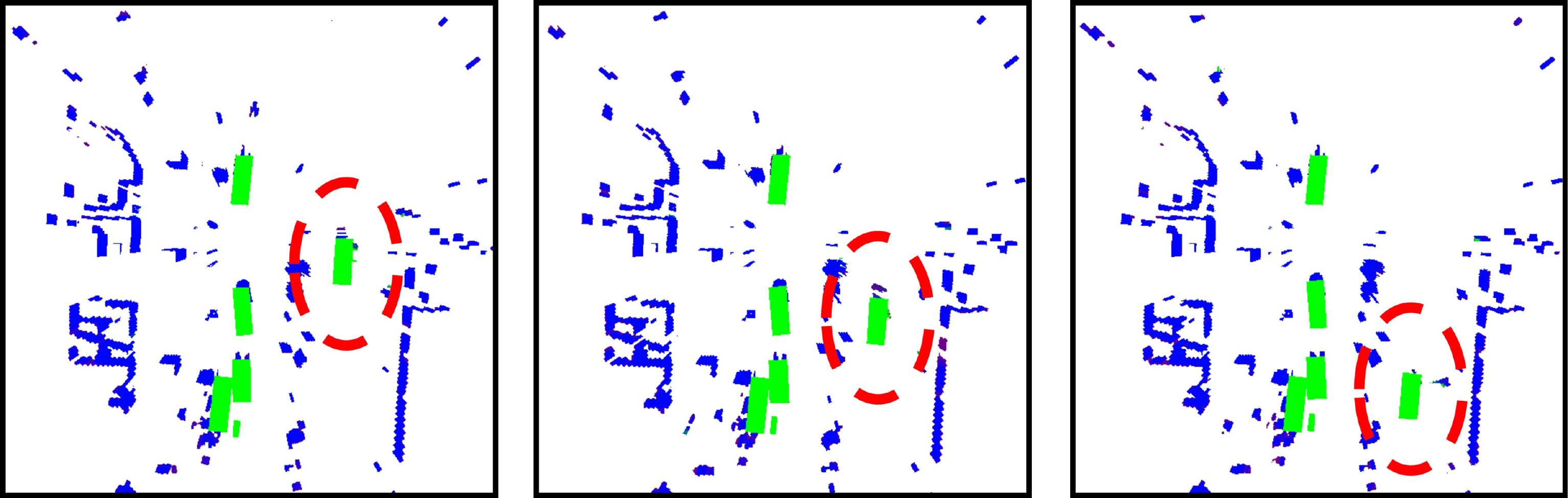}}
  \subfigure[PredRNN predictions]{
    \includegraphics[width=.475\textwidth]{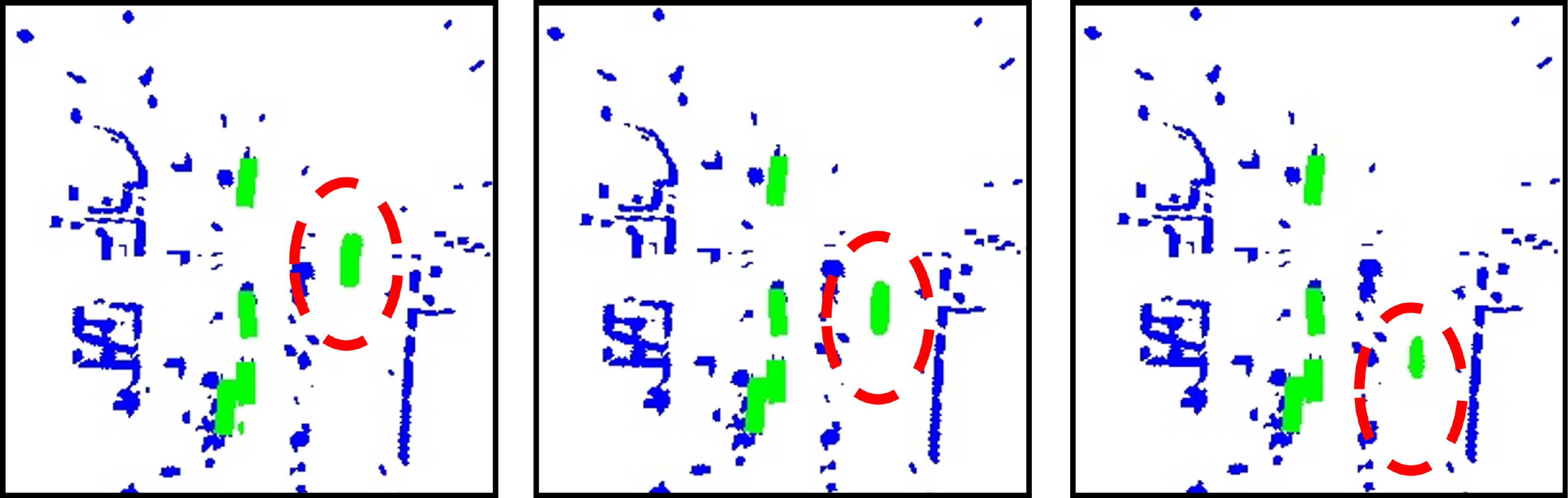}}
  \subfigure[ConvLSTM predictions]{
    \includegraphics[width=.475\textwidth]{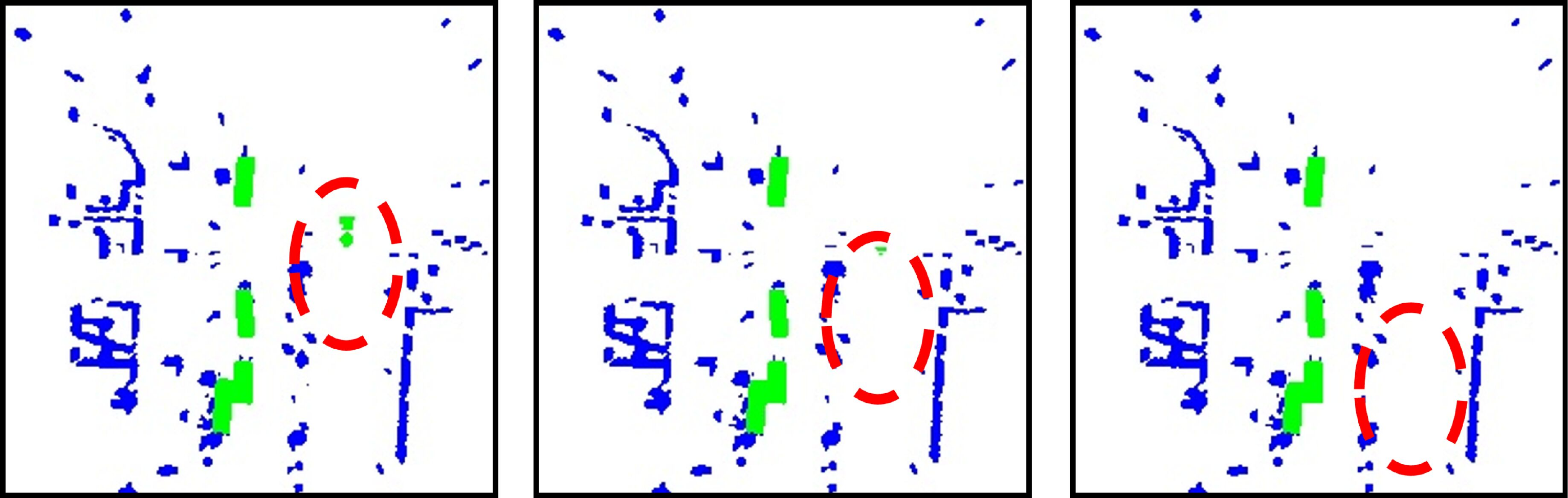}}
    \centering
    \caption{A scene depicting few static vehicles and a vehicle going in reverse direction. Comparison between two spatio-temporal learning networks over the future predictions of 1 sec, 2 sec and 3 sec.} 
	\label{fig:reverse_motion}
\end{figure}

\begin{figure}[!t]
  \subfigure[Ground truth ]{
    \includegraphics[width=.475\textwidth]{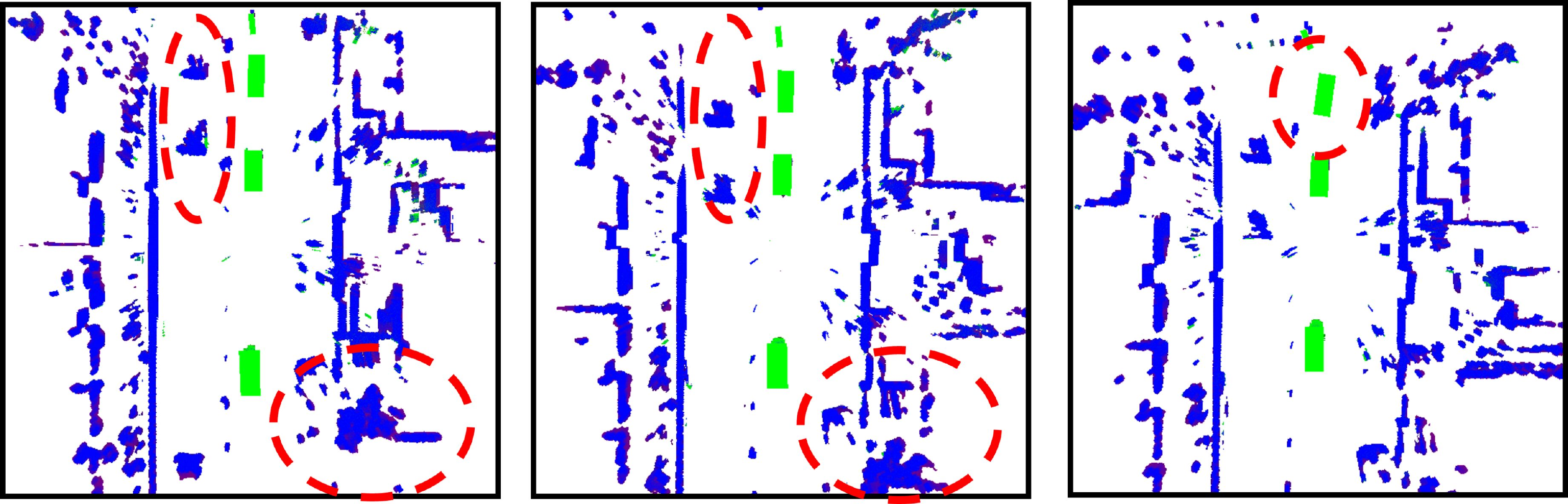}}
  \subfigure[PredRNN predictions]{
    \includegraphics[width=.475\textwidth]{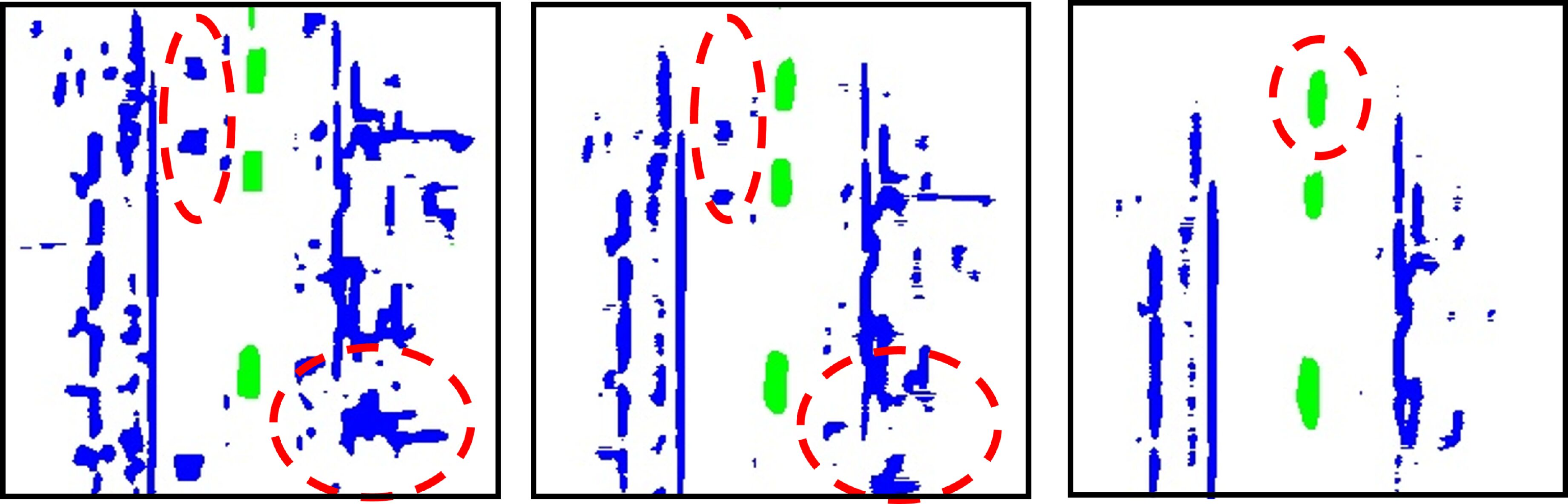}}
  \subfigure[ConvLSTM predictions]{
    \includegraphics[width=.475\textwidth]{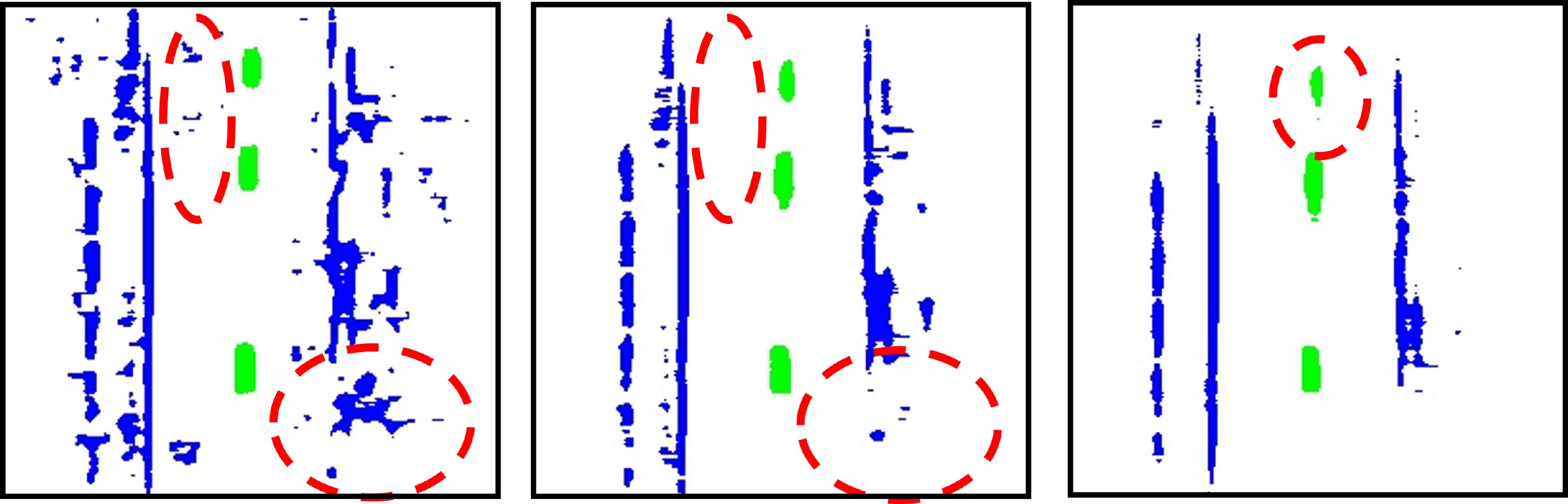}}
    \centering
    \caption{A scene describing three vehicles moving in forward direction along with the static environment. Predictions with spatio-temporal networks are presented over the future time horizon of 3 sec.} 
	\label{fig:static_obj}
\end{figure}

We consider a threshold of P$_{o}$ $>$ 0.6 for classifying the cells as occupied or not occupied. We use the same threshold for the vehicles and the rest of the environment predictions. The output of the model is a black and white image with a value of either 0 or 1 to indicate an empty or occupied state in the OGM. For representative purposes, in figure \ref{fig:reverse_motion} and \ref{fig:static_obj}, the cells with vehicle objects are represented by 'green' colour while static environment and objects other than vehicles are assigned `blue' colour.

Figure \ref{fig:reverse_motion} illustrates a sample driving scenario with few static vehicles and a vehicle moving in reverse direction on a straight road. The future predictions from our proposed pipeline using PredRNN and ConvLSTM are compared. The top row depicts ground truth OGMs at specific prediction times. The ego vehicle is fixed at the center and is traveling in the forward direction in the OGMs. The motion of the entire scene is relative to the ego vehicle. The cells with the vehicles are depicted in green and the rest of the static environment with other agents is represented in blue. In this scenario, four vehicles are static and there is a lone vehicle moving in the reverse direction with respect to the ego-vehicle. The reverse-moving vehicle is marked with dashed red lines for comparison across models. In figure \ref{fig:reverse_motion}(c), it can be seen that the network with ConvLSTM as backbone fails to predict the vehicle even 2 sec into the future whereas the PredRNN based network shows the vehicle in the environment for the entire 3 sec prediction horizon. However, it is to be noted that at 3 sec the predicted position of the vehicle lags in comparison to the actual position shown in the ground truth. This could be attributed to the high speed of the vehicle and a bias in the dataset with a large number of forward-moving vehicles and their motion. 

Figure \ref{fig:static_obj} is an example scenario of a double-lane road with static objects detected in the middle of the road. Apart from correctly predicting vehicles, it is crucial that the model has an accurate representation of free space in the drivable area. In figure \ref{fig:static_obj}(b), it can be seen that the PredRNN-based network can retain those small obstacles in the drivable area (circled with dashed red lines) up to 2 sec into the future, while ConvLSTM network fail to retain these crucial smaller things. Also, for multiple vehicles moving in the forward direction, the prediction from the ConvLSTM model gets blurry and smaller for the vehicle at the front towards the end of 3 sec as highlighted in the figure. 

Certain scenarios where the ego-vehicle turns and the entire scene rotates in the OGM are difficult to model and this would require further studies where the static environment of the future scenes or stored environment map with static objects can be provided to the networks. Certain situations are inherently difficult to account for such as a new vehicle appearing in the scene for a few frames before $t=0$ or a vehicle tuning scenario at an intersection. However, in certain situations, using the information from the structure of free space, the network was able to correctly predict a vehicle's turning. 

\begin{figure*}[t]

  \subfigure[]{
    \includegraphics[width=.33\textwidth]{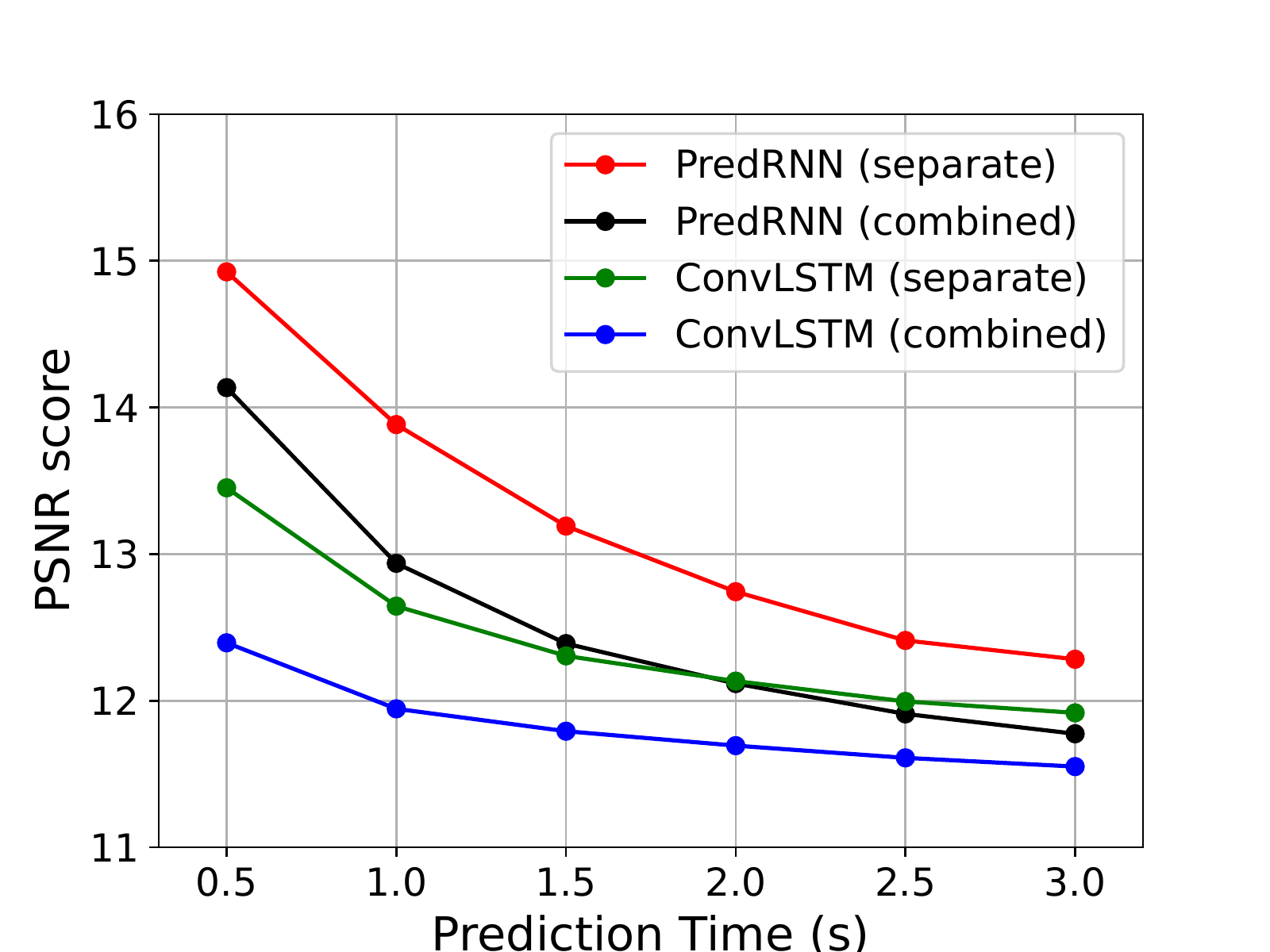}}
  \subfigure[]{
    \includegraphics[width=.33\textwidth]{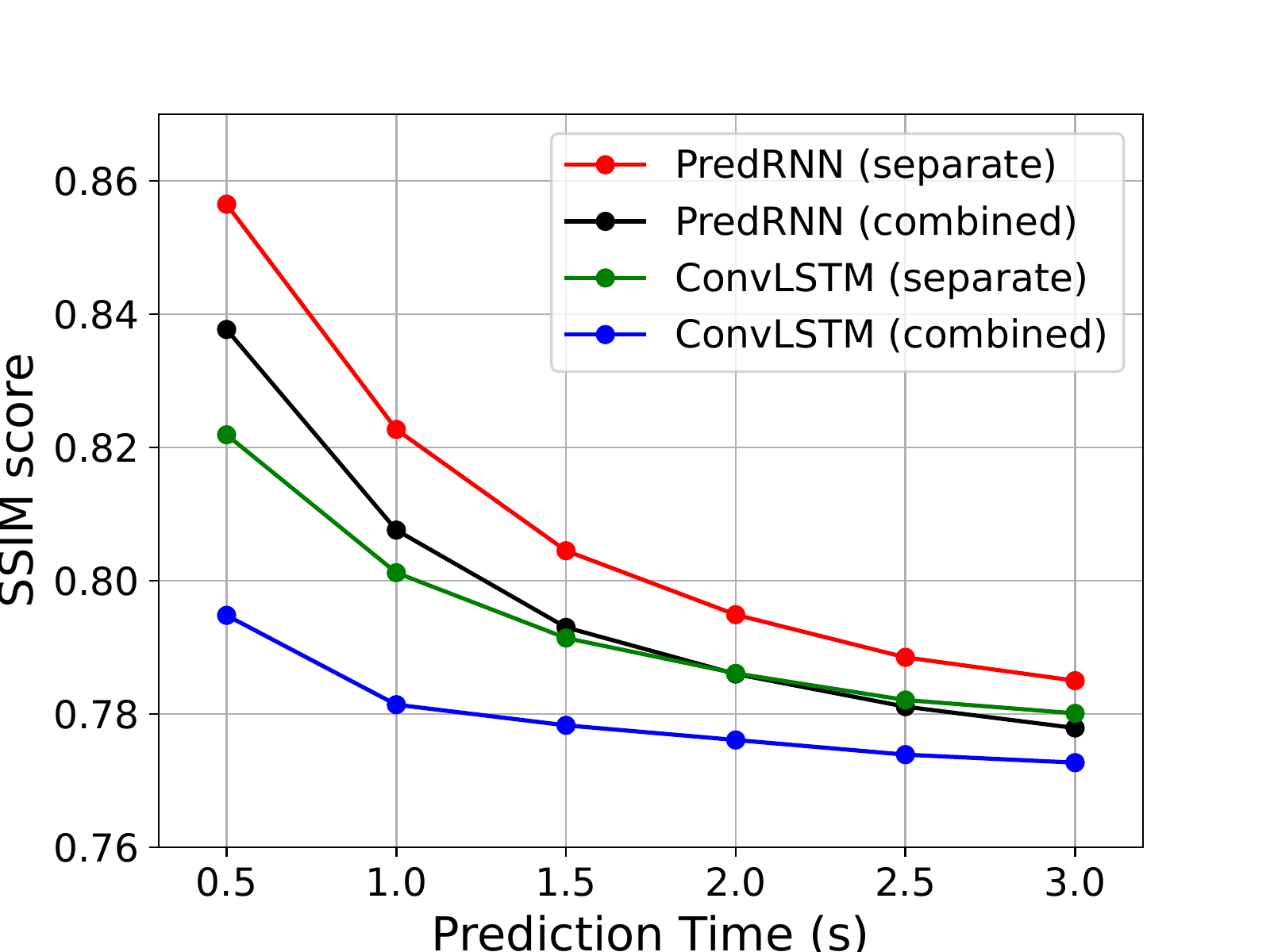}}  
   \subfigure[ ]{
    \includegraphics[width=.33\textwidth]{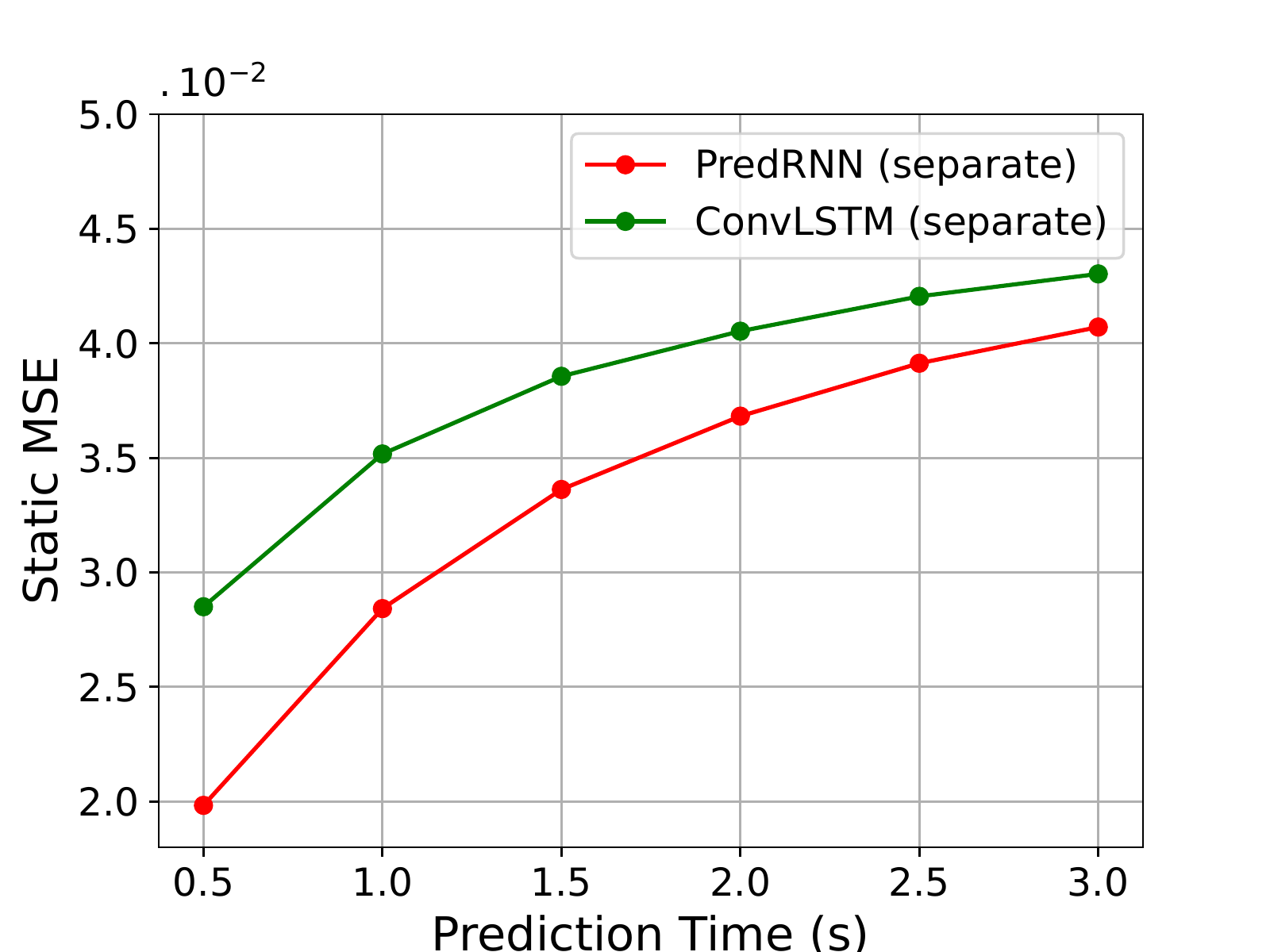}}
  \subfigure[]{
    \includegraphics[width=.33\textwidth]{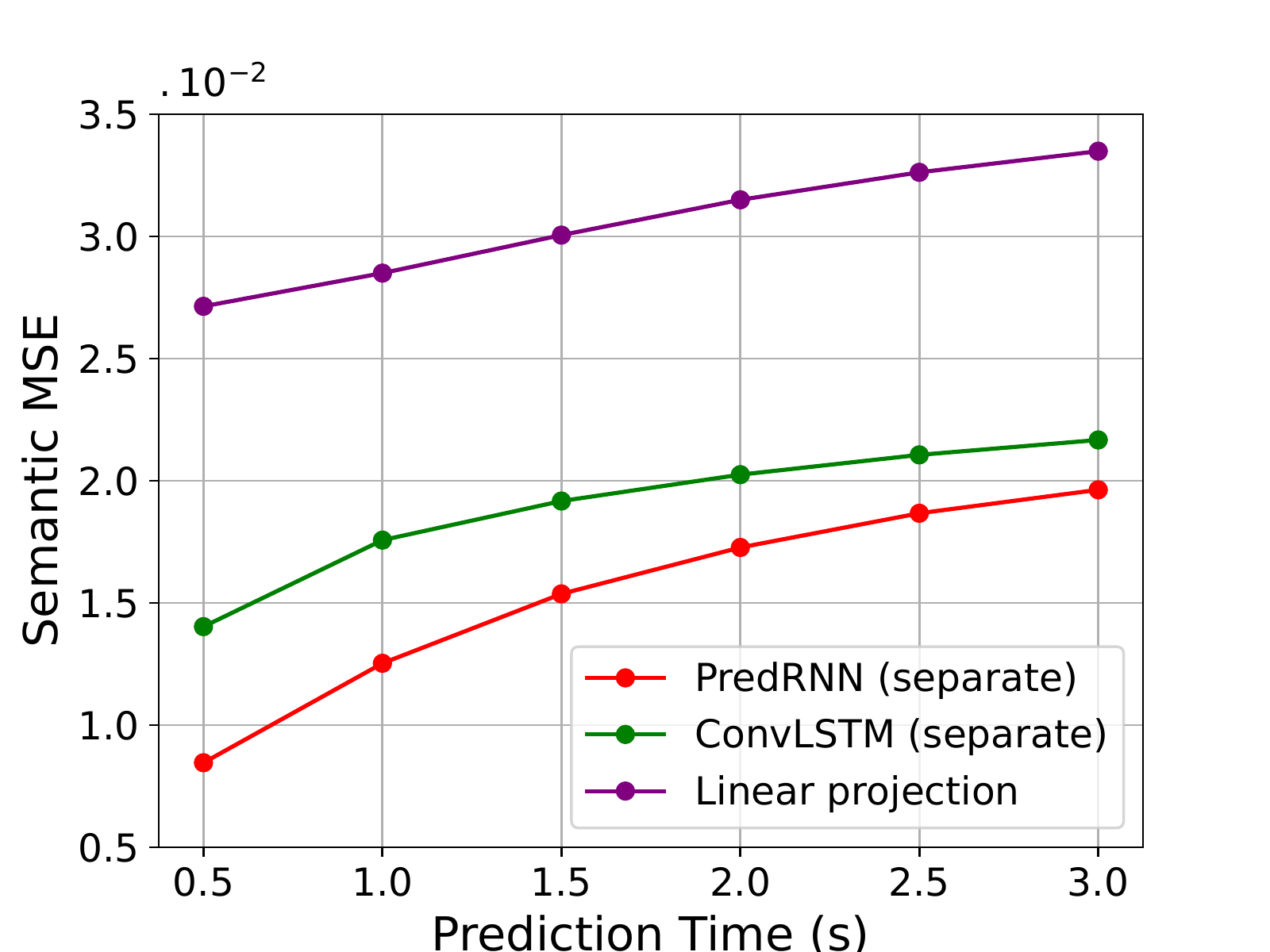} }
    \centering
    \caption{Frame-wise PSNR($\uparrow$), SSIM($\uparrow$), Static MSE ($\downarrow$), and Semantic MSE ($\downarrow$) results on the generated OGM dataset. The prediction horizon is 3 sec during training and testing phases. Note that PredRNN (combined) and ConvLSTM (combined) predict an entire OGM, thus separate Static and Semantic MSE cannot be reported for these cases. The Semantic MSE result from the linear projection of vehicle's bounding boxes is also presented in (d). } 
	\label{fig:evaluation_metrics}
\end{figure*}

\subsection{Quantitative results}
We quantitatively measure the prediction performance of the networks between the ground truth and predicted frames based on metrics like F1 score, Peak Signal-to-Noise Ratio (PSNR), Structural Similarity Index Measure (SSIM) \cite{wang2004image}, and Mean Squared Error (MSE). For each specific future time instance, we report mean values of 1200 test scenarios. These scenarios represent various urban traffic conditions and navigation challenges.

\begin{table}[t]
\setlength{\tabcolsep}{4.5pt}
\begin{tabularx}{1\linewidth}{c c c c}
 & & Prediction time \\
 \hline
Trained networks & \textbf{1 sec} & \textbf{2 sec}  & \textbf{3 sec}\\
\hline
PredRNN {\scriptsize (separate)} &  85.32 $\pm$ 4.31  & 83.87 $\pm$ 4.87 & 82.92 $\pm$ 4.92 \\
PredRNN {\scriptsize (combined)} &  84.21 $\pm$ 4.26  & 82.84 $\pm$ 4.69 & 81.61 $\pm$ 5.04 \\ 
ConvLSTM {\scriptsize (separate)} & 82.39 $\pm$ 4.85 &	81.83 $\pm$ 5.00 &	81.07 $\pm$ 5.01\\
ConvLSTM {\scriptsize (combined)} & 81.44 $\pm$ 4.96 &	81.27 $\pm$ 5.02 & 80.62 $\pm$ 5.00\\

\hline
\end{tabularx}
\caption{Mean F1 scores (\%) corresponding to 1200 test scenarios subjected to individual time frames.}
\label{table:F1_scores}
\end{table}

Table \ref{table:F1_scores} shows the F1 score comparisons between two variants of PredRNN and ConvLSTM models over the future prediction for 3 secs. Here, separate prediction imply separately predicting static and semantic objects as described in the model architecture (Figure \ref{fig:model_architecture}). While combined prediction implies predicting the entire OGM as shown in figure \ref{fig:teaser}. It is evident that for both models, separate prediction yields slightly better results as compared to combined prediction. Moreover, PredRNN performs better than ConvLSTM across all time frames. As expected, a performance drop is observed with increasing prediction steps but still, the obtained scores are comparable to environment prediction with static ego-vehicle\cite{schreiber2019long}.

Figures \ref{fig:evaluation_metrics}(a) and (b) present the OGM prediction plots of PSNR and SSIM for two networks up to future 3 secs respectively. While PSNR estimates the pixel-level similarity between two images, SSIM measures the perceived changes in structural information for the pixel that are spatially close to each other. We set the sliding window size as 11 for SSIM evaluation. PredRNN outperforms ConvLSTM implying that the predicted frames from the former are more structurally similar compared to later. Also, for each of these networks the separate prediction scheme is better than combined predictions. As we report the mean of several test scenarios, a trend in performance drop is visible with prediction time. This is addressed to the fact that certain scenarios wherein the ego vehicle suddenly makes turns are difficult to predict and reconstruct at longer time horizons. 

Figures \ref{fig:evaluation_metrics}(c) and (d) show the computed pixel-level MSE for grid cells corresponding to static and semantic objects separately. MSE values are expected to increase with time due to the accumulation of prediction errors. At each time step, PredRNN outperforms the ConvLSTM and linear projection methods, exhibiting performance consistency as observed in Figures \ref{fig:reverse_motion} and \ref{fig:static_obj}. In general, the metrics across different datasets cannot be compared. However, the scale of MSE metric is comparable. Accordingly, the scale of both static and semantic MSE is in the range of $10^{-2}$, both are comparable to the previous literature \cite{itkina2019dynamic,toyungyernsub2020double}. This range is attributed to the fact that MSE is calculated over the entire image dominated by free space which the model is able to predict easily. 

\section{Conclusion and future work}
We presented a pipeline for producing long-time (future 3 secs) OGM predictions of the urban environment in the case of dynamic ego-vehicle. These predictions were done separately for static and semantic obstacles with spatio-temporal networks. Accordingly, the networks were able to forecast the position of dynamic objects under different traffic scenarios and overcome the challenge of vanishing surrounding objects. We also showed that networks perform slightly better while separately predicting the static and semantic objects as compared to their combined predictions. 

Our OGMs are based on nuScenes dataset and include several real-world urban traffic conditions. We believe that it is vital for current techniques in AVs to model and predict these kinds of real-world scenarios and thus, we publicly release this dataset for further enhancement in OGM based research domain. For the future work, the current pipeline can be extended for performing multi-modal predictions.

\section{Acknowledgement}
This research work has been supported by the French Government in the scope of the FUI STAR and ES3CAP projects. Authors would like to thank Unmesh Patil for conducting experiments with linear projection model.

\bibliographystyle{IEEEtran}
\bibliography{IEEEabrv,Bibliography}

\vfill

\end{document}